\documentclass[10pt,twocolumn,letterpaper]{article}

\usepackage[pagenumbers]{cvpr} 

\usepackage{graphicx}
\usepackage{amsmath}
\usepackage{amssymb}
\usepackage{booktabs}
\usepackage{enumerate}
\usepackage{multirow}
\usepackage[accsupp]{axessibility} 

\usepackage[pagebackref,breaklinks,colorlinks]{hyperref}
\usepackage[capitalize]{cleveref}
\crefname{section}{Sec.}{Secs.}
\Crefname{section}{Section}{Sections}
\Crefname{table}{Table}{Tables}
\crefname{table}{Tab.}{Tabs.}

\def\cvprPaperID{12250} 
\def\confName{CVPR}
\def\confYear{2023}

\begin{document}

\title{Seeing Through the Glass: Neural 3D Reconstruction of Object Inside a Transparent Container}

\author{Jinguang Tong$^{1,2}$, Sundaram Muthu$^{2}$, Fahira Afzal Maken$^{2}$, Chuong Nguyen$^{1,2}$, Hongdong Li$^{1}$ \\ 
$^{1}$The Australian National University \quad $^{2}$Data61, CSIRO \\ 
{\tt\small \{jinguang.tong, chuong.nguyen, hongdong.li\}@anu.edu.au} \\ 
{\tt\small \{sundaram.muthu, fahira.afzalmaken, chuong.nguyen\}@data61.csiro.au}
}
\maketitle

\begin{abstract}

In this paper, we define a new problem of recovering the 3D geometry of an object confined in a transparent enclosure. 
We also propose a novel method for solving this challenging problem. 
Transparent enclosures pose challenges of multiple light reflections and refractions at the interface between different propagation media \eg air or glass. 
These multiple reflections and refractions cause serious image distortions which invalidate the single viewpoint assumption. 
Hence the 3D geometry of such objects cannot be reliably reconstructed using existing methods, such as traditional structure from motion or modern neural reconstruction methods. 
We solve this problem by explicitly modeling the scene as two distinct sub-spaces, inside and outside the transparent enclosure. 
We use an existing neural reconstruction method (NeuS) that implicitly represents the geometry and appearance of the inner subspace. 
In order to account for complex light interactions, we develop a hybrid rendering strategy that combines volume rendering with ray tracing. 
We then recover the underlying geometry and appearance of the model by minimizing the difference between the real and rendered images. 
We evaluate our method on both synthetic and real data. Experiment results show that our method outperforms the state-of-the-art (SOTA) methods. 
Codes and data will be available at \url{https://github.com/hirotong/ReNeuS}

\end{abstract}
    

\begin{figure}[ht]
\setlength{\belowcaptionskip}{0pt}
    \centering
    \begin{minipage}{0.95\linewidth}
    \begin{subfigure}[b]{1\linewidth}
    \centering
    \begin{minipage}{0.5\linewidth}
        \includegraphics[width=\linewidth, height=\linewidth]{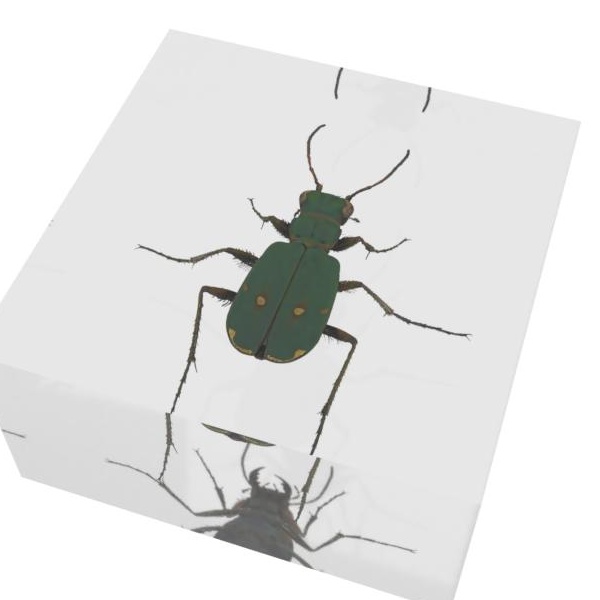}
        \caption{Reference Image}
        \label{fig:figure1a}
    \end{minipage}%
    \begin{minipage}{0.5\linewidth}
        \includegraphics[width=\linewidth,height=\linewidth]{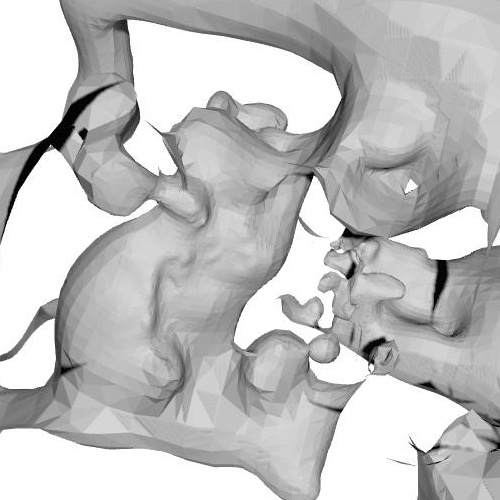}
        \caption{COLMAP \cite{Schonberger2016StructureFromMotionRevisited}}
    \end{minipage}
    \caption*{}
    \end{subfigure}
    \begin{subfigure}[b]{1\linewidth}
    \centering
    \begin{minipage}{0.5\linewidth}
        \includegraphics[width=\linewidth, height=\linewidth]{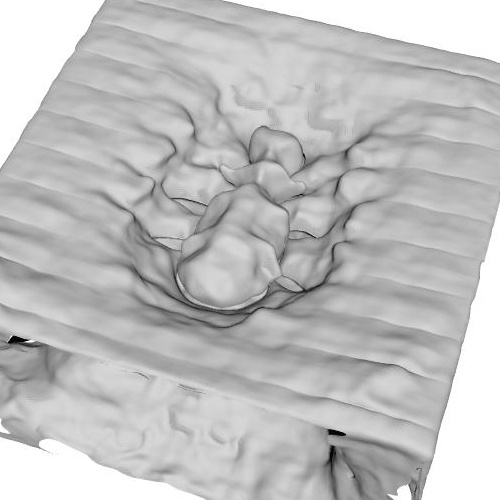}
        \caption{NeuS \cite{wang2021NeuSLearningNeural}}
    \end{minipage}%
    \begin{minipage}{0.5\linewidth}
        \includegraphics[width=\linewidth,height=\linewidth]{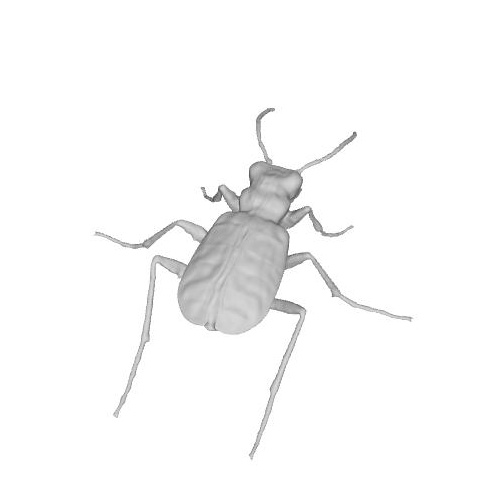}
        \caption{\textbf{Ours}}
    \end{minipage}
    \caption*{}
    \end{subfigure}
    \end{minipage}
    \caption{We propose a new research problem of recovering shape of objects inside a transparent container as shown in (a) and solve it by a physically-based neural reconstruction solution. Other methods (b) and (c) without consideration of the optical effects fail to reconstruct, (d) our proposed ReNeuS can retrieve high-quality mesh of the insect object.}
    \label{fig:fig1}
    \vspace{-2em}
\end{figure}

\section{Introduction}
\label{sec:intro}

Digitization of museum collections is an emerging topic of increasing interest, especially with the advancement of virtual reality and Metaverse technology.
Many famous museums around the world have been building up their own digital collections 
\footnote{\url{https://collections.louvre.fr/en/}}
\footnote{\url{https://www.metmuseum.org/}}
for online exhibitions.
Among these collections, there is a kind of special but important collections such as insects \cite{DigitalCollections,csiroAustralianNationalInsect}, human tissues \cite{MuseumHumanDisease}, aquatic creatures \cite{csiroAustralianNationalFish} and other fragile specimens that need to be preserved inside some hard but transparent materials (\eg resin, glass) as shown in \cref{fig:figure1a}. 
In order to digitize such collections, we abstract a distinct research problem which is seeing through the glassy outside and recovering the geometry of the object inside. 
To be specific, the task is to reconstruct the 3D geometry of an object of interest from multiple 2D views of the scene. 
The object of interest is usually contained in a transparent enclosure, typically a cuboid or some regular-shaped container. 

A major challenge of this task is the severe image distortion caused by repeated light reflection and refraction at the interface of air and transparent blocks.  
Since it invalidates the single viewpoint assumption \cite{glaeser2000reflections}, most existing multi-view 3D reconstruction methods tend to fail. 

One of the most relevant research problems is 3D reconstruction through refractive interfaces, known as Refractive Structure-from-Motion (RSfM) \cite{agrawal2012TheoryMultilayerFlat,chadebecq2017RefractiveStructureFromMotionFlat}. 
In RSfM, one or more parallel interfaces between different propagation media, with different refractive indices, separate the optical system from the scene to be captured. 
Light passing through the optical system refracts only once at each interface causing image distortion. 
Unlike RSfM methods, we deal with intersecting interfaces where multiple reflections and refractions take place. 
Another related research topic is reconstruction of transparent objects \cite{han2018DenseReconstructionTransparent, li2020LookingGlassNeurala, lyu2020DifferentiableRefractiontracingMesh}. 
In this problem, lights only get refracted twice while entering and exiting the target.
In addition to the multiple refractions considered by these methods, our method also tackles the problem of multiple reflections within the transparent enclosure. 

Recently, neural implicit representation methods \cite{michalkiewicz2019ImplicitSurfaceRepresentations, mildenhall2020nerf,sitzmann2020ImplicitNeuralRepresentations, sun2022Neural3DReconstructiona, wang2021NeuSLearningNeural} achieve state-of-the-art performance in the task of novel view synthesis and 3D reconstruction showing promising ability to encode the appearance and geometry. 
However, these methods do not consider reflections. 
Although NeRFReN \cite{guo2022NeRFReNNeuralRadiance} attempts to incorporate the reflection effect, it cannot handle more complicated scenarios with multiple refraction and reflection. 

In order to handle such complicated cases and make the problem tractable, we make the following two reasonable simplifications:

\begin{enumerate}[i.]
    \item \label{simp:known_geometry} \textbf{Known geometry and pose of the transparent block.} 
    It is not the focus of this paper to estimate the pose of the interface plane since it is either known \cite{agrawal2012TheoryMultilayerFlat, chadebecq2017RefractiveStructureFromMotionFlat} or calculated as a separate research problem \cite{Hu2021AbsoluteRelativePose} as is typically assumed in RSfM.
    
    \item \label{simp:homo} \textbf{Homogeneous background and ambient lighting.} Since the appearance of transparent objects is highly affected by the background pattern, several transparent object reconstruction methods \cite{han2018DenseReconstructionTransparent, lyu2020DifferentiableRefractiontracingMesh} obtain ray correspondence with a coded background pattern. 
    To concentrate on recovering the shape of the target object, we further assume monochromatic background with only homogeneous ambient lighting conditions similar to a photography studio.
\end{enumerate}

To handle both \textbf{Re}flection and \textbf{Re}fraction in 3D reconstruction, we propose \textbf{ReNeuS}, a novel extension of a neural implicit representation method NeuS \cite{wang2021NeuSLearningNeural}. 
In lieu of dealing with the whole scene together, a divide-and-conquer strategy is employed by explicitly segmenting the scene into internal space, which is the transparent container containing the object, and an external space which is the surrounding air space.
For the internal space, we use NeuS \cite{wang2021NeuSLearningNeural} to encode only the geometric and appearance information. 
For the external space, we assume a homogeneous background with fixed ambient lighting as described in simplification \ref{simp:homo}. 
In particular, we use a novel hybrid rendering strategy that combines ray tracing with volume rendering techniques to process the light interactions across the two sub-spaces. 
The main contributions of this paper are as follows: 
\begin{itemize}
    \item We introduce a new research problem of 3D reconstruction of an object confined in a transparent enclosure, a rectangular box in this paper. 
    \item We develop a new 3D reconstruction method, called ReNeuS, that can handle multiple light reflections and refractions at intersecting interfaces. 
    ReNeuS employs a novel hybrid rendering strategy to combine the light intensities from the segmented inner and outer sub-spaces.
    \item We evaluate our method on both synthetic as well as a newly captured dataset that contains $10$ real insect specimens enclosed in a transparent glass container.  
\end{itemize}

\begin{figure*}[t]
    \setlength{\belowcaptionskip}{0pt}
    \centering
    \includegraphics[width=0.95\linewidth]{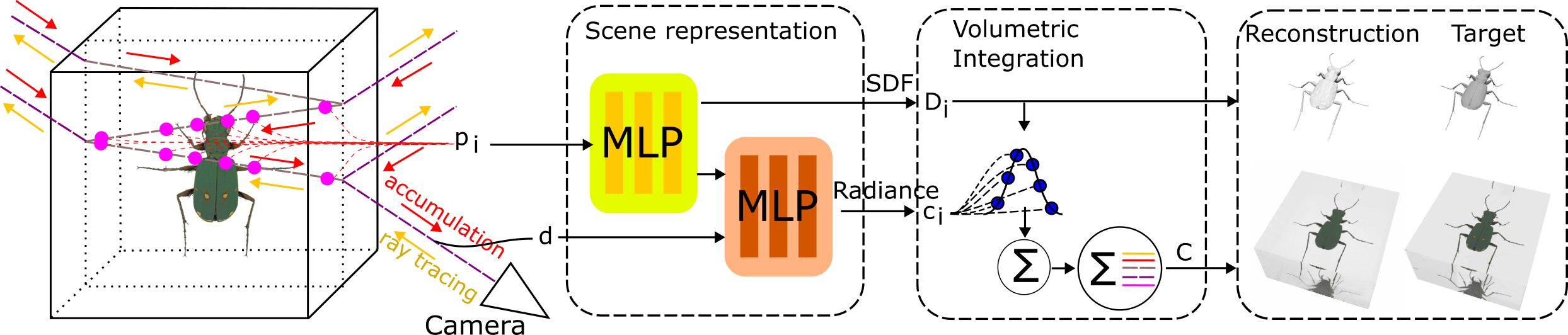}
    \caption{An overview of the ReNeuS framework. The scene is separated into two sub-spaces \wrt the interface. 
    The internal scene is represented by two multi-layer perceptrons (MLPs) for both geometry and appearance. 
    For neural rendering, we recursively trace a ray through the scene and collect a list of sub-rays. 
    Our neural implicit representation makes the tracing process controllable. 
    We show a ray tracing process with the depth of recursion $D_{re} = 3$ here. 
    Color accumulation is conducted on irradiance of the sub-rays in the inverse direction of ray tracing. 
    We optimized the network by the difference between rendered image and the ground truth image. 
    Target mesh can be extracted from SDF by Marching Cubes \cite{lorensen1987MarchingCubesHigha}}.
    \label{fig:overview}
    \vspace{-2em}
\end{figure*}

\section{Related work}
\label{sec:related_work}
\textbf{Refractive Structure from Motion}
The Refractive Structure-from-Motion (RSfM) problem involves recovering 3D scene geometry from underwater images where light refraction at the interface between different propagation media causes distortions which invalidates the assumption of a single viewpoint \cite{ Glaeser2000ReflectionsOR} (Pinhole camera assumption fails). 
The refractive structure from motion method developed in \cite{jordt-sedlazeck2013RefractiveStructurefromMotionUnderwater} accurately reconstructs underwater scenes that are captured by cameras confined in an underwater glass housing. 
This is achieved by incorporating multiple refractions in the camera model that takes place at interfaces of varying densities of glass, air, and water.  
\cite{chadebecq2017RefractiveStructureFromMotionFlat} presents a robust RSfM framework to estimate camera motion and 3D reconstruction using cameras with thin glass interfaces that seal the optical system. 
To achieve this, \cite{chadebecq2017RefractiveStructureFromMotionFlat} considers the refractive interface explicitly by proposing a refractive fundamental matrix. 
Unlike the methods described in \cite{chadebecq2017RefractiveStructureFromMotionFlat,  jordt-sedlazeck2013RefractiveStructurefromMotionUnderwater}, where the camera is confined in front of a piece of glass, our method works with small objects, such as insects, that are embedded within a glassy box.

\textbf{Transparent Object Reconstruction}
Acquiring the 3D geometry of transparent objects is challenging as general-purpose scanning and reconstruction techniques cannot handle specular light transport phenomena. 
3D shape reconstruction of highly specular and transparent objects is studied under controlled settings in \cite{wu2018Full3DReconstructiona}. 
The method in \cite{wu2018Full3DReconstructiona} mounts the object on a turntable. 
The model is optimized using the refraction normal consistency, silhouette consistency, and surface smoothness. 
However, this method requires a known refractive index assumption of the transparent object. 
\cite{han2018DenseReconstructionTransparent} presents a method for reconstruction of transparent objects with unknown refractive index. 
This method partially immerses the object in a liquid to alter the incident light path. Then the object surface is recovered through reconstructing and triangulating such incident light paths. 
\cite{ lyu2020DifferentiableRefractiontracingMesh} captures fine geometric details of the transparent objects by using correspondences between the camera view rays and the locations on the static grey-coded background. 
Recently,\cite{li2020LookingGlassNeurala} proposes a neural network to perform 3D reconstruction of transparent objects using a small number of mobile camera images. 
This is achieved by modelling the reflections and refractions in a rendering layer and a cost volume refinement layer. 
The rendering layer optimizes the surface normal of the object to guide the point cloud reconstruction.

\textbf{Neural 3D reconstruction}
The use of neural networks has been widely used in recent years to predict new views and geometric representations of 3D scenes and objects based on a limited set of images captured with known camera poses. 
Several works exist that represent 3D structures using deep implicit representations. 
The main advantage over using conventional voxel grids or meshes is the fact that the implicit representations are continuous and memory efficient to model shape parts of the objects. 
\cite{michalkiewicz2019ImplicitSurfaceRepresentations} uses implicit representations of curves and surfaces of any arbitrary topology using level sets. 
\cite{mildenhall2020nerf} presents NeRF that synthesizes novel views of complex scenes by optimizing a continuous neural radiance field representation of a scene from a set of input images with known camera poses. 
\cite{guo2022NeRFReNNeuralRadiance} presents NeRFReN, which is based on NeRF and allows us to model scenes with reflections. 
In particular, \cite{guo2022NeRFReNNeuralRadiance} splits the scene into transmitted and reflected components and models the two components with separate neural radiance fields. 
Michael \etal propose UNISURF \cite{oechsle2021UNISURFUnifyingNeural}, a unified framework that couples the benefits of neural implicit surface models (for surface rendering)  with those of neural radiance fields (for volume rendering) to reconstruct accurate surfaces from multi-view images.
However, this method is limited to representing solid and non-transparent surfaces. 
The method presented in \cite{yariv2020MultiviewNeuralSurface} relaxes the requirement of multi-view camera calibration by describing an end-to-end neural system that is capable of learning 3D geometry, appearance, and cameras from masked 2D images and noisy camera initialization. 
NeuS \cite{wang2021NeuSLearningNeural} performs high-quality reconstruction of objects with severe occlusions and complex geometries by combining radiance fields and occupancy representations. 
This method integrates both neural volumetric representation of NeRF and neural implicit representations using SDFs to represent the 3D surfaces. 
The implicit representation greatly improves the NeRF. 
However, NeuS cannot handle the severe image distortion created due to multiple reflections and refractions of the object placed inside a glass. 
Hence, our method ReNeuS modifies the NeuS \cite{wang2021NeuSLearningNeural} to overcome the problem by using a hybrid rendering strategy.

\section{Method}
\label{sec:method}
Given a set of $N$ posed images and a known transparent rectangular box geometry, we aim to recover the geometry of the internal object. 
As shown in \cref{fig:overview}, we proposed a neural reconstruction method called ReNeuS which extends NeuS to work on scenes with complex light refractions and reflections.
To decompose such a complicated scene, we first divide it into two sub-spaces (inside and outside).
Following NeuS \cite{wang2021NeuSLearningNeural}, the internal space is represented by an implicit neural network which contains an implicit surface representation with signed distance function (SDF) and a volumetric appearance representation.
To render a color image of the scene, we explicitly trace its ray path through the scene.
For all the sub-rays in the ray path, we retrieve the radiance of sub-rays according to their location (inner or outer) and then physically accumulate them to get the color of a ray. 
We call this a hybrid rendering method.
All of the network parameters are then optimized by minimizing the difference between the rendered image and the ground truth image.



\subsection{Neural Implicit Surface Revisited}
\label{sec:neus}
\textbf{Scene representation.} 
Neural Implicit Surface (NeuS) \cite{wang2021NeuSLearningNeural} represents scenes with neural networks which contain a neural surface representation of signed distance function (SDF) and a volumetric radiance representation. 
Two independent multi-layer perceptrons (MLP) are used to encode the scene's geometry and appearance.
To be specific, the geometric MLP: $g_\theta: (\mathbf{x}) \rightarrow (\mathbf{s})$ with $\theta$ as parameters, represent the scene as an SDF field that it takes over a point coordinate $\mathbf{x}$ and outputs its signed distance value $\mathbf{s}$.
From $g(\mathbf{x})$, the surface of the object $\mathcal{S}$ can be extracted from a zero-level set of its SDF as:
\begin{equation} \label{eqn:sdf}
    \mathcal{S}=\left\{\mathbf{x}\in\mathbb{R}^3\ |\ \mathit{g}_{\theta}(\mathbf{x})=0\right\}
\end{equation}
Since the appearance is view-dependent, another appearance MLP: $f_\phi: (\mathbf{x},\mathbf{v},\mathbf{n}) \rightarrow (\mathbf{c})$ encodes radiance information of a point $\mathbf{c}$ with its position $\mathbf{x}$, view direction $\mathbf{v}$ and the normal vector $\mathbf{n}$.

\textbf{Volume Rendering}
Let's first recall the volume rendering equation. 
Given a light ray $\boldsymbol\ell(t) = \mathbf{o} + t\mathbf{d}$, the color $\mathbf{C}(\boldsymbol\ell)$ is calculated by the weighted integration of radiance along the ray by:
\begin{equation} \label{eqn:volume_rendering}
    \mathbf{\hat{C}}(\boldsymbol\ell)=\int_{t_n}^{t_f} w(t) c(\mathbf{x}(t), \mathbf{d}) \mathrm{d} t
\end{equation}
where $w(t)$ is the weight at point $\mathbf{x}(t)$ and $c(\mathbf{x}(t), \mathbf{d})$ is the radiance emitted from point $\mathbf{x}(t)$ along the direction $\mathbf{d}$. 
In neural radiance field based representation \cite{mildenhall2020nerf}, the weight is a composition of volume density $\sigma(\mathbf{x}(t))$ and accumulated transmittance $T(t)$:
\begin{align}
    w(t) &= T(t) \sigma(\mathbf{x}(t)), \label{eqn:nerf_weight} \\ 
    \text{where} \  T(t) &= \mathrm{exp}\left(-\int_{t_n}^{t} \sigma(\mathbf{x}(s)) \mathrm{d}s \right) \label{eqn:nerf_transmittance}
\end{align} 
In order to apply volume rendering on the SDF network, Peng \etal \cite{wang2021NeuSLearningNeural} derives an \emph{unbiased} and \emph{occlusion-aware} weight function as:
\begin{equation} \label{eqn:neus_weight}
    w(t) = T(t) \rho(t)
\end{equation}
with the aid of an opaque density function:
\begin{equation} \label{eqn:odf}
    \rho(t) =\max\left(\frac{-\frac{\mathrm{d} \Phi_s}{\mathrm{~d} t}(g(\mathbf{x}(t)))}{\Phi_s(g(\mathbf{x}(t)))}, 0\right)
\end{equation}
where $\Phi_s(x) = (1 + e^{-sx})^{-1}$ is the Sigmoid function.
Finally, the integration in \cref{eqn:volume_rendering} is estimated by using numerical quadrature \cite{max1995OpticalModelsDirecta} with hierarchical sampling strategy. 
For more detail, refer to the original papers \cite{mildenhall2020nerf,wang2021NeuSLearningNeural}.

\subsection{ReNeuS Scene Representation} 
\label{sec:scene_representation}
Since the problem origins from the heterogeneity of the scene, our solution is to separate the scene by the interface of different mediums (\eg air/glass in our case), and obtain two sub-spaces \emph{internal} $\mathbb{S}_{in}$ and \emph{external} $\mathbb{S}_{out}$.

\textbf{Internal space $\mathbb{S}_{in}$}.
We adapt NeuS to represent the local internal space $\mathbb{S}_{in}$ which contains the target object.
As described in \cref{sec:neus}, the appearance of $\mathbb{S}_{in}$ is represented by a volumetric radiance field.
From the principle of volumetric radiance field \cite{max1995OpticalModelsDirecta}, light interaction (including absorbing, reflecting \etc) is explained by the probability of interaction and has been embedded in the volume rendering equation.
In our case, a light ray can go straightly through $\mathbb{S}_{in}$ and its color can be retrieved by volume rendering as \cref{eqn:volume_rendering}. 

\textbf{External space $\mathbb{S}_{out}$}.
As our simplification \ref{simp:homo}, the external space is assumed to be empty with homogeneous ambient lighting $\mathbf{C}^{out} = \mathbf{C}^{ambient}$.


\subsection{ReNeuS Renderer}
\label{sec:renderer}
To render a color image of the scene, we utilize a novel rendering strategy that combines ray tracing \cite{parker2005InteractiveRayTracing} and volume rendering \cite{max1995OpticalModelsDirecta} techniques. 

\textbf{Ray tracing}. 
In computer graphics, ray tracing is the process of modelling light transport for rendering algorithms.
We slightly modified the ordinary ray tracing method.
In ReNeuS, the appearance of internal space $\mathbb{S}_{in}$ is represented as a volumetric radiance field.
When tracing a ray through $\mathbb{S}_{in}$, we do not try to explicitly model the interaction with the target object but let the ray go straightly through the space.
This allows us to get much better control of the scene and we only need to consider the light interactions (reflection and refraction) happening on the interfaces of different sub-spaces. 

To be specific, for a ray $\boldsymbol\ell_i$, we calculate the intersection of $\boldsymbol\ell_i$ and the transparent box. 
If there is no hit, the tracing process will be terminated with only the original ray $\boldsymbol\ell_i$ recorded.
If $\boldsymbol\ell_i$ hits the box, we explicitly trace the reflected ray $\boldsymbol\ell_i^r$ and refracted ray $\boldsymbol\ell_i^t$ according to the Law of Reflection and Snell's Law \cite{born2013principles}, so that we have:
\begin{equation} \label{eqn:ray_tracing}
    \boldsymbol\ell_i^r,\ \boldsymbol\ell_i^t = \mathcal{RE}(\boldsymbol\ell_i, \mathbf{n}_i)
\end{equation}
where $\mathbf{n}_i$ is the surface normal at the intersection and $\mathcal{RE}$ represents the operation of \textbf{re}flection and \textbf{re}fraction.
Besides, the reflectance $R_i$ is calculated according to the Fresnel equations under ``natural light'' assumption \cite{born2013principles} and also the transmittance $T_i^{re} = 1 - R_i$. 
Note the transmittance $T_i^{re}$ is different from $T(t)$ in \cref{eqn:nerf_transmittance} that $T_i^{re}$ is the transmittance of refraction light while $T(t)$ represents the transmittance of the space along ray direction.
The color of reflected and refracted rays can be expressed as:
\begin{gather}
    \mathbf{C}(\boldsymbol\ell_i^r) = R_i \cdot \hat{\mathbf{C}}(\boldsymbol\ell_i^r)  \label{eqn:irradiance_reflection} \\
    \mathbf{C}(\boldsymbol\ell^i_t) = T_i^{re} \cdot \hat{\mathbf{C}}(\boldsymbol\ell_i^t) \label{eqn:irradiance_refraction}
\end{gather}
The tracing process is performed recursively on $\boldsymbol\ell_i^r$ and $\boldsymbol\ell_i^t$ and constrained by the depth of recursion $D_{re}$ which is a hyper-parameter in our method.
After ray tracing, each original camera ray $\boldsymbol\ell_i$ is extended to $\mathbb{L}_i$ which contains a set of traced rays. 
 
\textbf{Color Accumulation.} 
The color of a ray emitted from the camera $\boldsymbol\ell_i$ is restored by accumulating that among all the sub-rays in $\mathbb{L}_{i}$.
\begin{equation}
    \mathbf{C(\boldsymbol\ell_i)} = \underset{\mathbf{m}_i \in \mathbb{L}_{i}}{\mathcal{ACC}} (\mathbf{C}(\mathbf{m}_i))
\end{equation}
where $\mathcal{ACC}$ means physically-based color accumulation.
To do this, we first retrieve the color of each sub-ray $\mathbf{m}_i \in \mathbb{L}_i$.  
For internal rays, the color is obtained by volume rendering. 
And for external rays, a pre-defined background color $\mathbf{C}^{out}$ is directly assigned.
Next, the ray color is accumulated progressively in the reverse order of ray tracing. 
For a ray $\boldsymbol\ell$, if meets the interface, the color $\mathbf{C}(\boldsymbol\ell)$ is modified according to \cref{eqn:irradiance_reflection} and (\ref{eqn:irradiance_refraction}) and when passing through $S^{in}$, we further multiply the color $\mathbf{C}(\boldsymbol\ell)$ with the accumulated transmittance along the ray, computed as:
\begin{equation} \label{eqn:ray_transmittance}
    T_{\boldsymbol\ell} = \mathrm{exp}\left(-\int_{t_s}^{t_e} \rho(s) \mathrm{d}s \right)
\end{equation}
where $t_s$ and $t_e$ indicate the start and end of the ray.
 
\textbf{Working with Irradiance}.
Since we use a physical-based renderer, it should work with scene irradiance rather than image intensity.
The camera response function (CRF) describes the relationship between image intensity and scene irradiance.
To compensate for the nonlinear CRF, we assume that the appearance MLP predicts colors in linear space and adopts a gamma correction function to the final output:
\begin{equation}
    I = \mathcal{G}(\mathbf{C(\boldsymbol\ell)}),
\end{equation}
where $\mathcal{G}(\mathbf{C}) = \mathbf{C}^{\frac{1}{2.2}}$ is the gamma correction function.

\subsection{Loss Function}
Following previous works \cite{mildenhall2020nerf, yariv2020MultiviewNeuralSurface, wang2021NeuSLearningNeural}, the parameters of ReReuS are merely optimized and supervised by multi-view images with known poses.
A binary mask $M_{in}$ of the transparent enclosure is also used. 
To be specific, we randomly sample a batch of pixels. 
For each pixel $p$ in the batch, a camera ray $\boldsymbol\ell_p$ is generated with the corresponding camera pose. 
After that, we trace the camera ray $\boldsymbol\ell_p$ and get a ray set $\mathbb{L}_p$.
For each ray in $\mathbb{L}_p$, if it belongs to $\mathbb{S}_{in}$, we consistently sample $n$ points along the ray and do volume rendering. 
The color of the pixel $\mathbf{\hat{I}}_p$ is retrieved by accumulation and gamma correction as described in \cref{sec:renderer}. 

The overall loss function is given below, and described in detail afterwards:
\begin{equation} \label{eqn:loss_total}
    \mathcal{L} = \mathcal{L}_{color} + \lambda_1 \mathcal{L}_{trans} + \lambda_2 \mathcal{L}_{reg}
\end{equation}

\textbf{Photometric loss.} 
We empirically choose L1 loss between rendered pixel color and the ground truth as:
\begin{equation} \label{eqn:loss_rgb}
    \mathcal{L}_{color} = \frac{1}{|M_{in}|} \sum_{\mathbf{p} \in M_{in}}\left\|\hat{I}_p - I_p \right\|_1
\end{equation}

\textbf{Sparsity prior.}
Based on the crystalline appearance of the box, we exploit a sparsity prior that a considerable part of $\mathbb{S}_{in}$ (except the area occupied by the object) should be somehow transparent to the light. 
We apply this prior by regularizing the transmittance of $\mathbb{S}_{in}$ as:
\begin{equation} \label{eqn:sparsity_prior}
    \mathcal{L}_{trans} = \frac{1}{|M_{in}|} \sum_{p \in M_{in}} \sum_{\boldsymbol\ell \in \mathbb{L}_{p} \cap \mathbb{S}_{in}} \left\| 1 - T_{\boldsymbol\ell}\right\|
\end{equation}
This prior is valid in our case since the opaque object usually takes up a limited area of the transparent box. 
Experiment results show that $\mathcal{L}_{trans}$ helps to get cleaner geometry. 

\textbf{Regularization.} The same as NeuS, we apply the Eikonal regularization from \cite{Gropp2020ImplicitGeometricRegularizationa} on the sampled points to regularize the geometric MLP as:
\begin{equation} \label{eqn:eikonal}
    \mathcal{L}_{reg} = \mathbb{E}_{\mathbf{x}}\left(\left|\nabla_\mathbf{x} g_{\theta}(\mathbf{x})\right| - 1\right) ^ 2,
\end{equation}
where $\mathbf{x}$ represents all the sampled points on the rays inside $\mathbb{S}_{in}$ within a batch.

\section{Experiments}
\label{sec:experiment}

\begin{figure*}[ht]
\setlength{\abovecaptionskip}{0pt}
\setlength{\belowcaptionskip}{0pt}
    \centering
    \begin{minipage}{0.95\linewidth}
    \begin{subfigure}[b]{0.12\linewidth}
    \ \\
    \begin{minipage}{1\linewidth}
        \includegraphics[width=\linewidth, height=\linewidth]{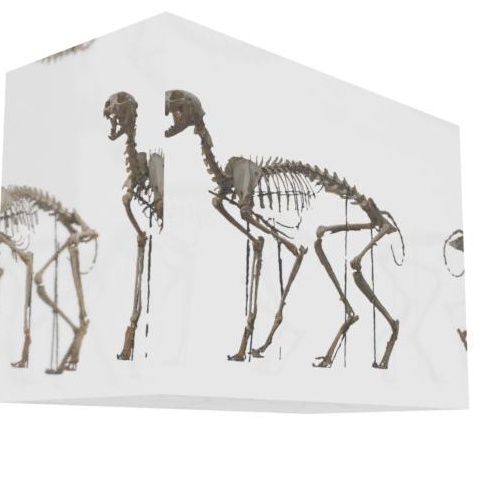}
        \includegraphics[width=\linewidth, height=\linewidth]{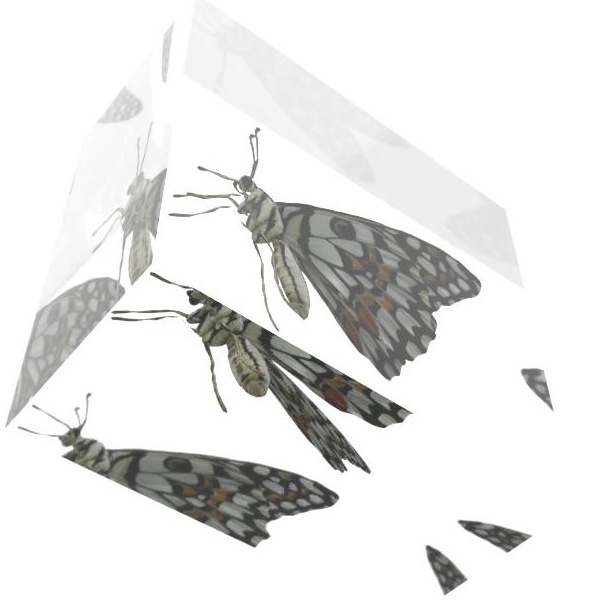}
        \includegraphics[width=\linewidth, height=\linewidth]{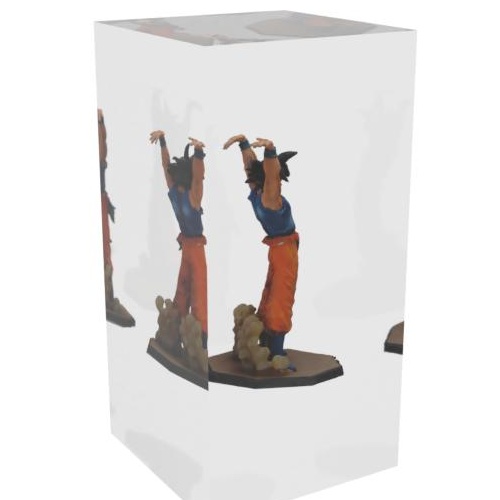}
        \includegraphics[width=\linewidth, height=\linewidth]{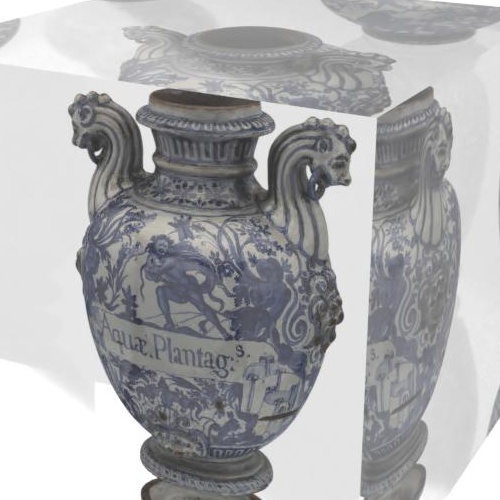}
        \centerline{\small GT view}
    \end{minipage}
    \caption*{}
    \end{subfigure}\vline\vspace{4pt}
    \begin{subfigure}[b]{0.36\linewidth}
    \centering \small \textbf{w/o box}
    \begin{minipage}{0.33\linewidth}
        \includegraphics[width=\linewidth, height=\linewidth]{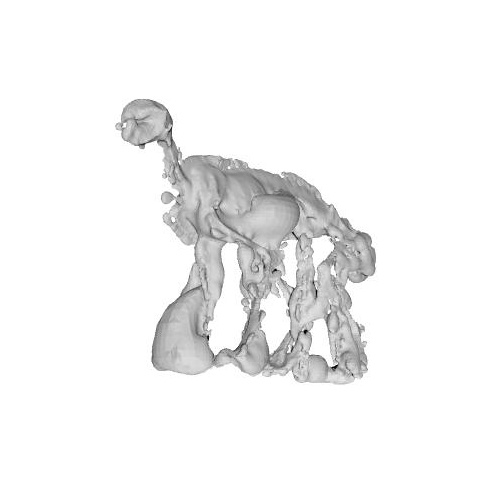}
        \includegraphics[width=\linewidth, height=\linewidth]{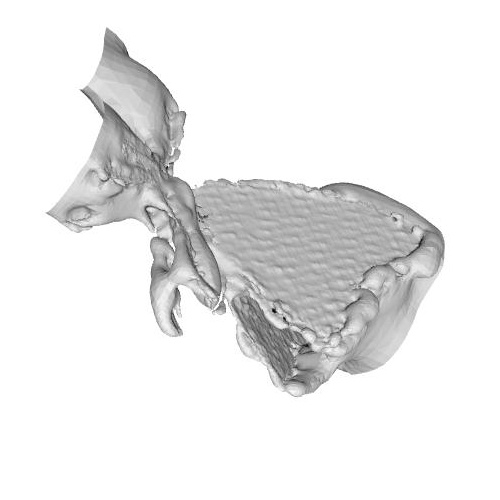}
        \includegraphics[width=\linewidth, height=\linewidth]{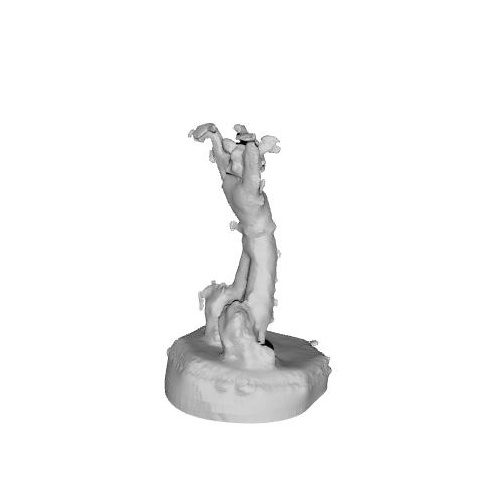}
        \includegraphics[width=\linewidth, height=\linewidth]{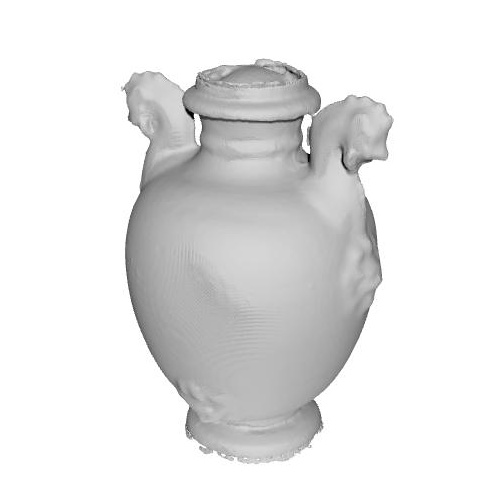}
        \centerline{\small $\text{COLMAP}_{\xi=0}$}
    \end{minipage}%
    \begin{minipage}{0.33\linewidth}
        \includegraphics[width=\linewidth, height=\linewidth]{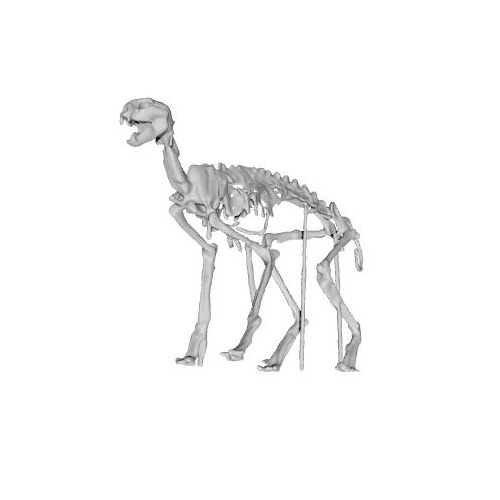}
        \includegraphics[width=\linewidth, height=\linewidth]{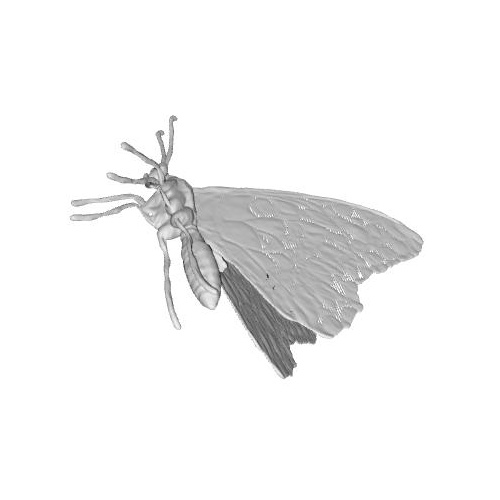}
        \includegraphics[width=\linewidth, height=\linewidth]{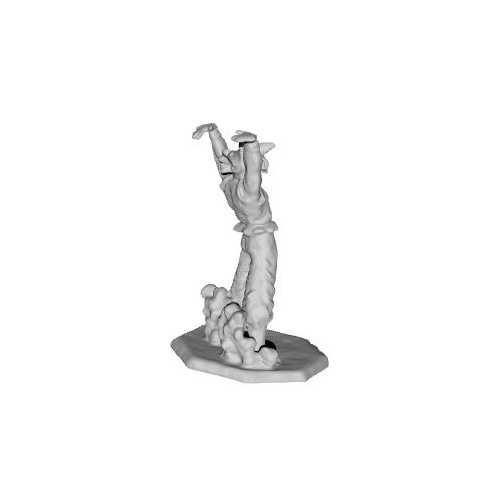}
        \includegraphics[width=\linewidth, height=\linewidth]{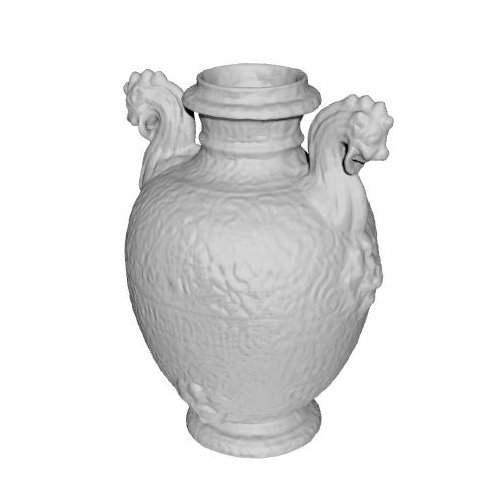}
        \centerline{\small IDR}
    \end{minipage}%
    \begin{minipage}{0.33\linewidth}
        \includegraphics[width=\linewidth, height=\linewidth]{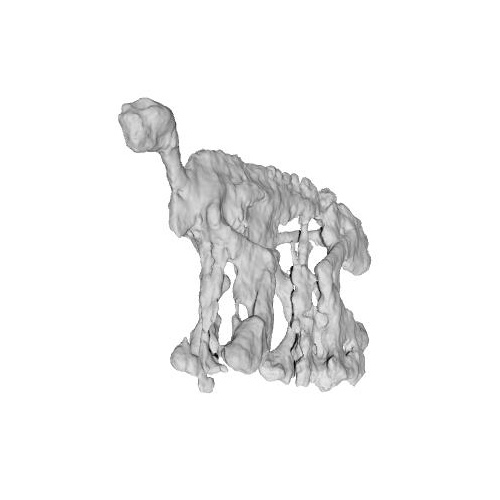}
        \includegraphics[width=\linewidth, height=\linewidth]{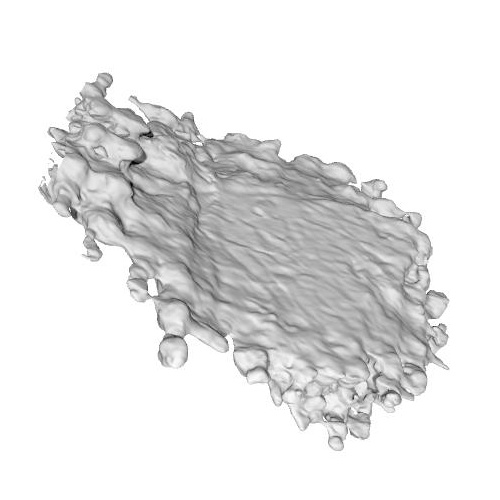}
        \includegraphics[width=\linewidth, height=\linewidth]{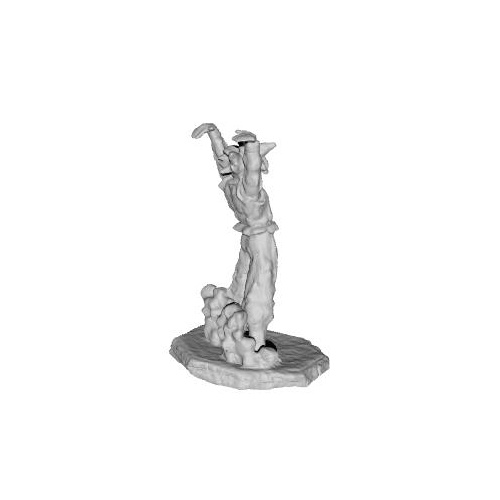}
        \includegraphics[width=\linewidth, height=\linewidth]{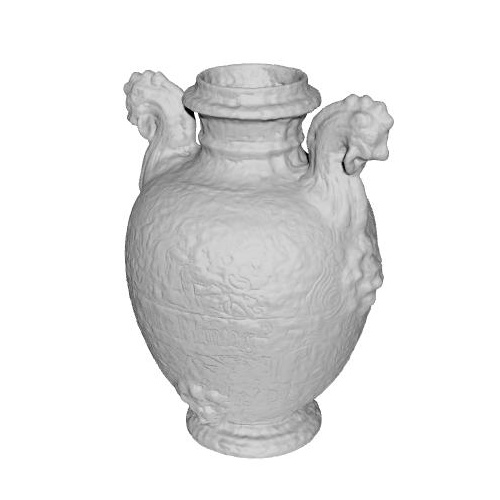}
        \centerline{\small NeuS}
    \end{minipage}
    \caption*{}
    \end{subfigure}\vline\vspace{4pt}
    \begin{subfigure}[b]{0.36\linewidth}
    \centering \small \textbf{w/box}
    \begin{minipage}{0.33\linewidth}
        \includegraphics[width=\linewidth, height=\linewidth]{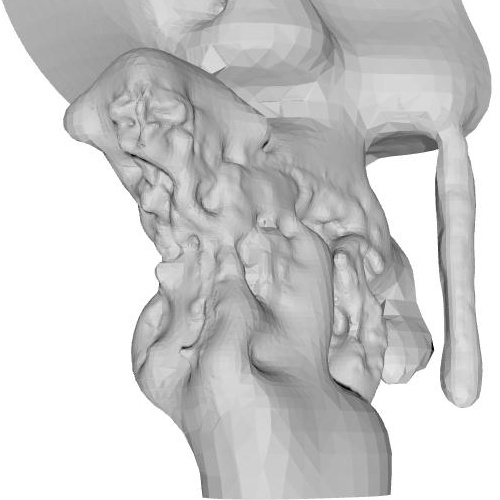}
        \includegraphics[width=\linewidth, height=\linewidth]{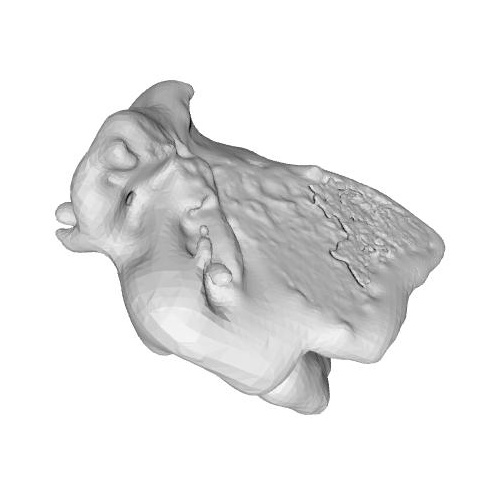}
        \includegraphics[width=\linewidth, height=\linewidth]{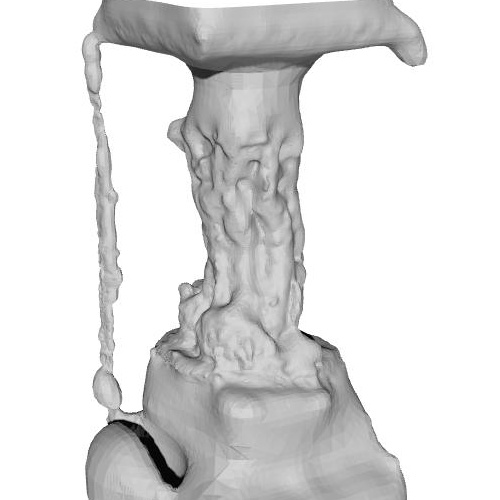}
        \includegraphics[width=\linewidth, height=\linewidth]{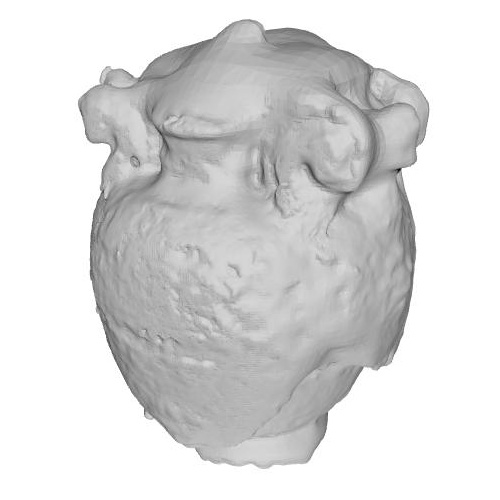}
        \centerline{\small $\text{COLMAP}_{\xi=0}$}
    \end{minipage}%
    \begin{minipage}{0.33\linewidth}
        \includegraphics[width=\linewidth, height=\linewidth]{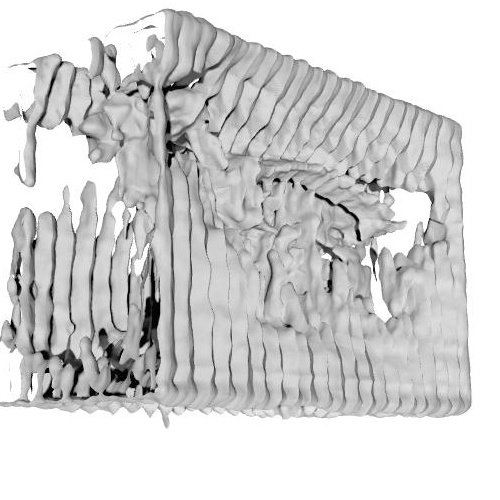}
        \includegraphics[width=\linewidth, height=\linewidth]{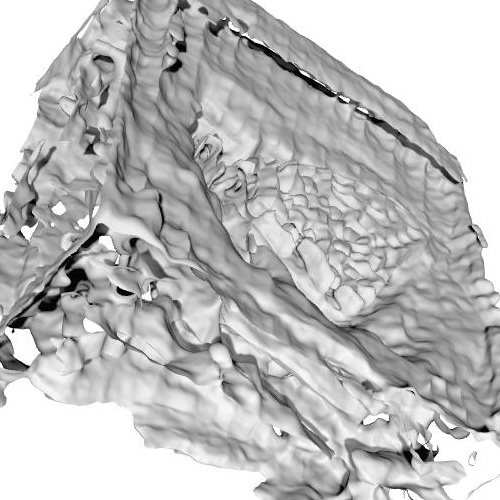}
        \includegraphics[width=\linewidth, height=\linewidth]{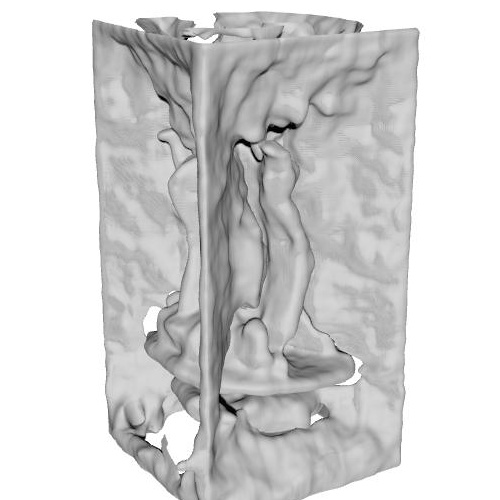}
        \includegraphics[width=\linewidth, height=\linewidth]{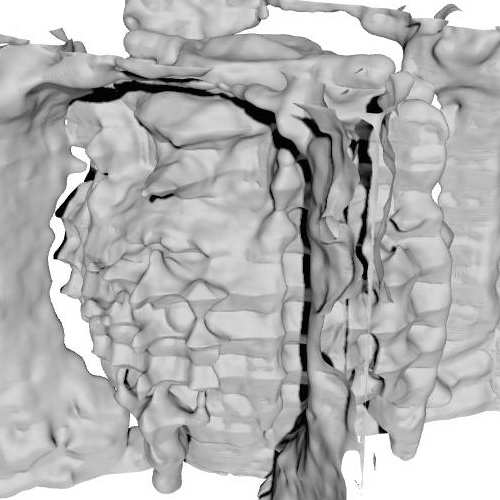}
        \centerline{\small NeuS}
    \end{minipage}%
    \begin{minipage}{0.33\linewidth}
        \includegraphics[width=\linewidth, height=\linewidth]{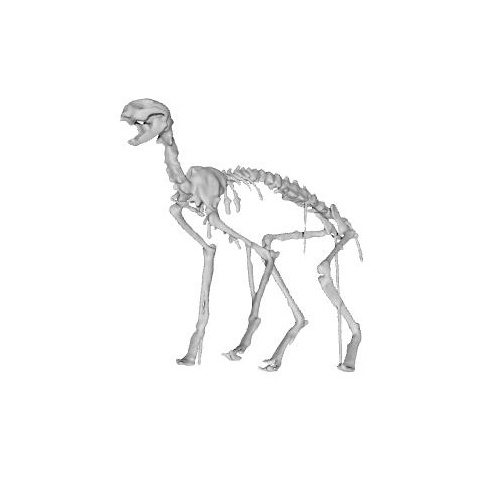}
        \includegraphics[width=\linewidth, height=\linewidth]{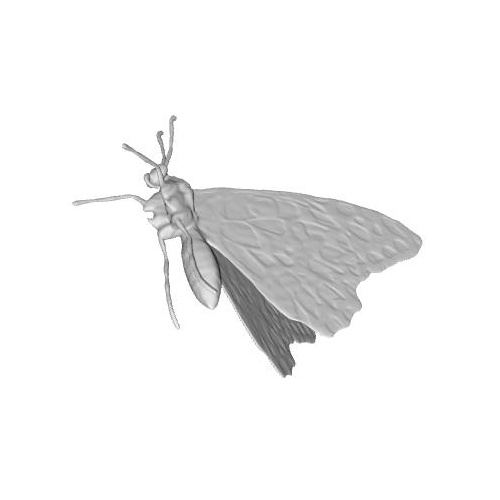}
        \includegraphics[width=\linewidth, height=\linewidth]{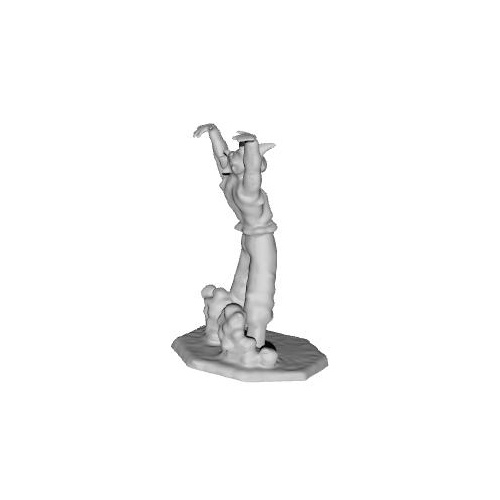}
        \includegraphics[width=\linewidth, height=\linewidth]{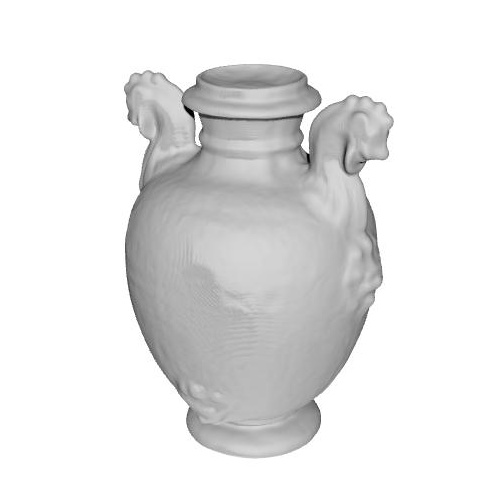}
        \centerline{\small \textbf{Ours}}
    \end{minipage}%
    \caption*{}
    \end{subfigure}\vline\vspace{4pt}
    \begin{subfigure}[b]{0.12\linewidth}
    \begin{minipage}{1\linewidth}
        \includegraphics[width=\linewidth, height=\linewidth]{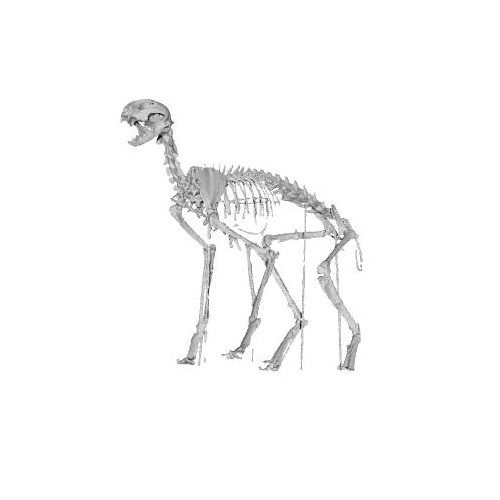}
        \includegraphics[width=\linewidth, height=\linewidth]{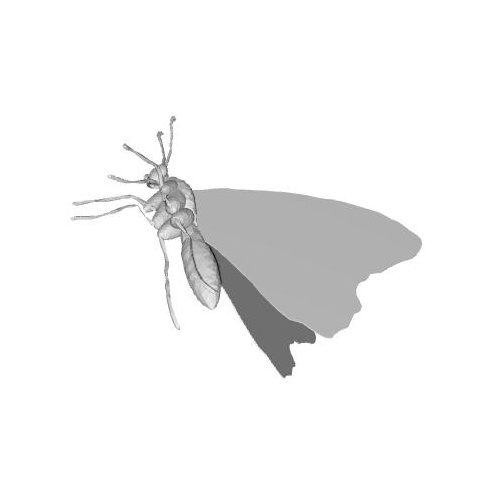}
        \includegraphics[width=\linewidth, height=\linewidth]{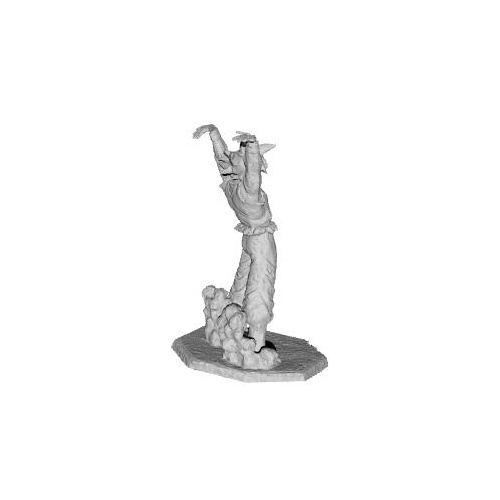}
        \includegraphics[width=\linewidth, height=\linewidth]{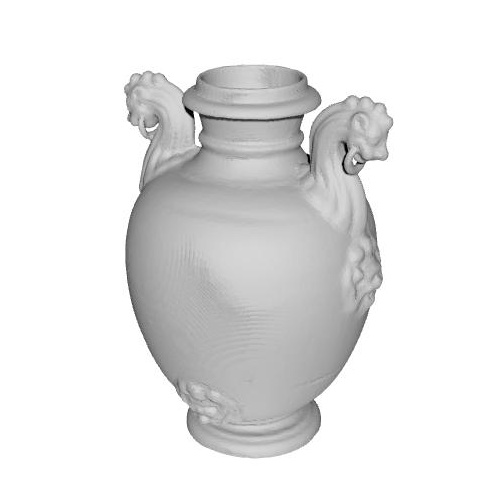}
        \centerline{\small GT mesh}
    \end{minipage}
    \caption*{}
    \end{subfigure}
    \end{minipage}
    \caption{Reconstruction of 3D shapes on the synthetic dataset. We qualitatively compare ReNeuS with baseline methods. Results show that our method successfully achieves high-quality reconstruction with transparent containers where other methods fail. }
    \label{fig:qualitative}
    \vspace{-1em}
\end{figure*}

\subsection{Dataset.}
We introduce two (one synthetic and another real) new datasets to evaluate our method and baseline methods since there is no existing dataset dedicated to our problem setting. 

\textbf{Synthetic Dataset.}
We choose 13 objects with complicated appearances and/or geometry to generate synthetic data.  
Blender \cite{Blender} with the physical-based render engine {\tt Cycles} is used to render photorealistic images. 
We put the object at the center of a box, manually adjust the size of the box to fit the object and then scale it within a unit sphere. 
The object generally has a Lambertian surface. 
The material of the box is adjusted by {\tt Principled BSDF} to appear as glass with a refractive index of 1.45. 
The camera pose is randomly sampled on a full sphere of radius 5.
For each camera pose, we not only render an image of the complete scene but also render another image without the transparent box. 
We render 60 viewpoints with a resolution of $800\times800$ for each scene, getting two subsets named w/ box and w/o box datasets.

\textbf{Real Dataset.} 
Another 10 real scenes of insect specimens data are captured. 
We capture the dataset in a photography light tent where the specimen is placed on a sheet of ChArUco board \footnote{\url{https://docs.opencv.org/4.x/df/d4a/tutorial_charuco_detection.html}} \cite{Garrido-Jurado2014AutomaticGenerationDetection} over a rotating display stand. 
The images are captured on a hemisphere with different azimuths and altitudes.
We use ChArUco board to calibrate the intrinsics as well as the pose of the camera.
As for the transparent box, we manually label it by edges in the images and calculate its dimensions and pose according to the known camera.
The transparent box mask is then generated by projecting the box to 2D images. 
Finally, we center-crop all images to be $1728 \times 1152$ pixels and randomly sample 60 images per scene. 
You can find more details about the dataset in the supplementary material. 

\subsection{Baseline methods.}
Since there is no specific method aiming at the same problem, we choose several state-of-the-art 3D reconstruction methods as baselines. 
To make our results more convincing, apart from the w/ box data, we also compare with the baselines on the w/o box dataset.

\noindent\textit{COLMAP} \cite{Schonberger2016StructureFromMotionRevisited, schonberger2016pixelwise} is a popular MVS method. 
For a fair comparison, we don't run structure-from-motion with COLMAP, but directly use the ground truth camera pose for reconstruction.
We obtain a mesh by applying screened Poisson Surface Reconstruction \cite{kazhdan2013screened} on the output point cloud of COLMAP.

\noindent\textit{IDR} \cite{yariv2020MultiviewNeuralSurface} is a neural rendering-based 3D reconstruction method that achieves results with high fidelity.
We run IDR with object mask of w/o box dataset only. 

\noindent\textit{NeuS.}
Our method is actually an extension of NeuS to an optically complex scene.
We compare with NeuS to verify our credit.

\begin{table*}[!hbtp]
	\begin{center}

	  {
	    \begin{tabular}{lcccc|cccr}
    		\toprule   
    		    
    	    \multirow{2}{*}{\raisebox{-\heavyrulewidth}{Items}} & \multicolumn{4}{c|}{\raisebox{-\heavyrulewidth}{{w/o box}}}  & \multicolumn{4}{c}{{w/ box}}\\
    	  
\cmidrule{2-5}
\cmidrule{6-9}

 & \small{COLMAP$_{\xi=0}$} & \small{COLMAP$_{\xi=7}$} &$IDR$&$ NeuS$& \small{COLMAP$_{\xi=0}$} & \small{COLMAP$_{\xi=7}$}&$ NeuS$&$\mathbf{ReNeuS}$\\ 
\cmidrule{1-9}

\multirow{1}{*}{beetle}
	
  &$1.10$	&$3.67$	&${\color{red} \mathbf{0.76}}$	&$2.42$ &$15.48$	&$17.55$	&$27.40$ &$\mathbf{1.40}$
 \\
 \multirow{1}{*}{box} 
  &$4.09$	&$5.28$	&${\color{red} \mathbf{0.51}}$	&$3.76$	&$35.38$	&$14.15$	&$15.43$	&$\mathbf{2.29}$
 \\
 \multirow{1}{*}{butterfly}
 
  &$6.10$	&$1.15$	&${\color{red} \mathbf{0.78}}$	&$9.98$&$12.44$	&$18.82$	&$19.85$	&$\mathbf{1.15}$ 
 \\
   \multirow{1}{*}{coral}
  &$4.93$	&$11.23$	&${\color{red} \mathbf{1.68}}$	&$2.11$ &$142$	&$17.32$	&$21.18$&$\mathbf{8.78}$
  \\
  \multirow{1}{*}{coral2}
  &$1.81$	&$2.75$	&$1.66$	&${\color{red} \mathbf{1.42}}$ &$21.38$	&$29.25$	&$20.75$	&$\mathbf{2.44}$
 \\

  \multirow{1}{*}{dinosaur}

    &$2.22$	&$1.18$	&${\color{red} \mathbf{0.82}}$	&$2.80$	&$28.15$	&$36.33$	&$15.46$	&$\mathbf{1.47}$
    \\
    \multirow{1}{*}{goku}
  
    &$2.21$	&$2.13$	&${\color{red} \mathbf{0.81}}$	&$1.37$ &$22.48$	&$11.09$	&$17.23$ &$\mathbf{1.48}$
   \\
    \multirow{1}{*}{insect}
    &$2.56$	&$0.96$	&${\color{red} \mathbf{0.62}}$	&$5.89$  &$45.78$	&$35.88$	&$21.71$&$\mathbf{1.10}$
    \\
    \multirow{1}{*}{insect2}
    &$1.34$	&$1.15$	&${\color{red} \mathbf{0.76}}$&$4.07$&$15.34$	&$18.65$	&$22.11$&$\mathbf{1.30}$
   \\
    \multirow{1}{*}{lobster}
    &$13.27$	&$12.20$	&$17.17$	&$12.97$&$40.52$	 	&$26.72$ &$14.61$&${\color{red} \mathbf{11.08}}$
   \\
    \multirow{1}{*}{shiba}
    &$2.33$	&$17.73$	&$2.24$	&${\color{red} \mathbf{2.22}}$&$24.30$	&$13.92$	&$20.30$&$\mathbf{3.38}$
   \\
    \multirow{1}{*}{statuette}
 
&$0.85$	&$0.92$	&$0.87$	&${\color{red} \mathbf{0.79}}$	  &$61.74$	&$22.16$	&$25.44$ &$\mathbf{2.26}$
\\
\multirow{1}{*}{vase}
&$1.37$	&${\color{red} \mathbf{1.30}}$	&$1.31$	&$1.35$	&$15.02$	&$15.06$	&$24.68$&$\mathbf{3.68}$
\\
\cmidrule{1-9}
\multirow{1}{*}{mean}
	&$3.22$	&$4.47$	&${\color{red} \mathbf{2.28}}$	&$3.72$&$35.86$	&$20.83$	&$20.59$	&$\mathbf{3.19}$
\\ \vspace{-0.75em} \\

		    \bottomrule 
	    \end{tabular}
	    }
	    
	\end{center}
\setlength{\abovecaptionskip}{0cm} 
\setlength{\belowcaptionskip}{0cm}
\caption{Qualitative evaluation on the synthetic dataset. We highlight the best result in terms of {\color{red} \textbf{w/o box}} and \textbf{w/ box} data respectively.}
\label{tab:double_column}
\vspace{-1em}
\end{table*}

\subsection{Implementation Details.}\label{sec:implement_details}
We implement ReNeuS on the top of NeuS \cite{wang2021NeuSLearningNeural}. 
We adopt the same network structure for geometric MLP $g_\theta$ and appearance MLP $f_\phi$ as NeuS.
The geometric MLP is initialized as \cite{Atzmon_2020_CVPR} to be an ellipsoid-like SDF field. 
Different from NeuS, we replace the activation function (Softplus) with a periodic activation function SIREN \cite{sitzmann2020ImplicitNeuralRepresentations} for its stronger capability for neural representation. 
The input parameters position $\mathbf{x}$ and view direction $\mathbf{v}$ are prompted with positional encoding \cite{mildenhall2020nerf}. 
The geometry of $\mathbb{S}_{out}$ is represented as a mesh of a regular box with known dimensions. 
For the real data, we manually calibrate it with the edges of the box.
Please refer to the supplementary material for more details.
For ReNeuS rendering, we set the recursion depth $D_{re} = 2$ as a trade-off between accuracy and efficiency.
The ambient lighting $C^{out}$ is set to $[0.8, 0.8, 0.8]$ (before gamma correction) for synthetic data, and we calibrate that by measuring the average background for real data. 
For loss function, \Cref{eqn:loss_total}, we empirically set $\lambda_1 = 0.1\ \text{and}\ \lambda_2 = 0.1$ for all experiments.
We sample 1024 rays per batch during training, with hierarchical sample \cite{mildenhall2020nerf}, totally 128 points (64 for coarse and 64 for fine) are sampled along a ray. 
No point outside is additionally sampled. 
We train all the models for totally 200k iterations on a single NVIDIA Tesla P100 GPU.
During evaluation time, we extract a mesh from the zero-level set of the geometric MLP $g_\theta$ with Marching Cubes \cite{lorensen1987MarchingCubesHigha}. 
Automatic post-processing is applied by clustering and retaining the mesh with most vertices.

\subsection{Evaluation on the synthetic dataset.}
We evaluate ReNeuS and other baseline methods on the synthetic dataset. 
We only run ReNeuS on the w/ box dataset while all the others are evaluated on both w/ box and w/o box datasets. 
For quantitative comparison, we report the Chamfer-$L_1$ distance between ground truth and the reconstruction. 
Due to the unbalanced number of points in the ground truth mesh, we uniformly sample $N = \max\left(10000,\ 0.2num_v\right)$ points from the reconstructed mesh as input, where $num_v$ is the number of vertices of the ground truth mesh. 
Note that for all results reported in the paper, we magnify it by $100\mathrm{x}$ for better comparison.

We report the experiment results in \cref{tab:double_column} and \cref{fig:qualitative}.
For the w/ box dataset, it's clear that ReNeuS outperforms other baselines by a large margin both quantitatively and qualitatively.
Comparing with the baselines on the w/o box dataset further proves the power of our method.
Even though working on much more challenging data, our method achieves comparable performance with the baselines methods, only falling behind IDR \cite{yariv2020MultiviewNeuralSurface} which needs a ground truth mask for training. 
It's worth noting that ReNeuS get superior performance than NeuS, even on data without a container box. 
We assume this to be the contribution of reflection. 
Every time when a ray is reflected, it's like we have a new view of the scene which helps with reconstruction.

\begin{figure*}[ht]
\setlength{\abovecaptionskip}{0pt}
\setlength{\belowcaptionskip}{0pt}
    \centering
    \begin{subfigure}[b]{0.23\linewidth}
    \includegraphics[width=1\linewidth]{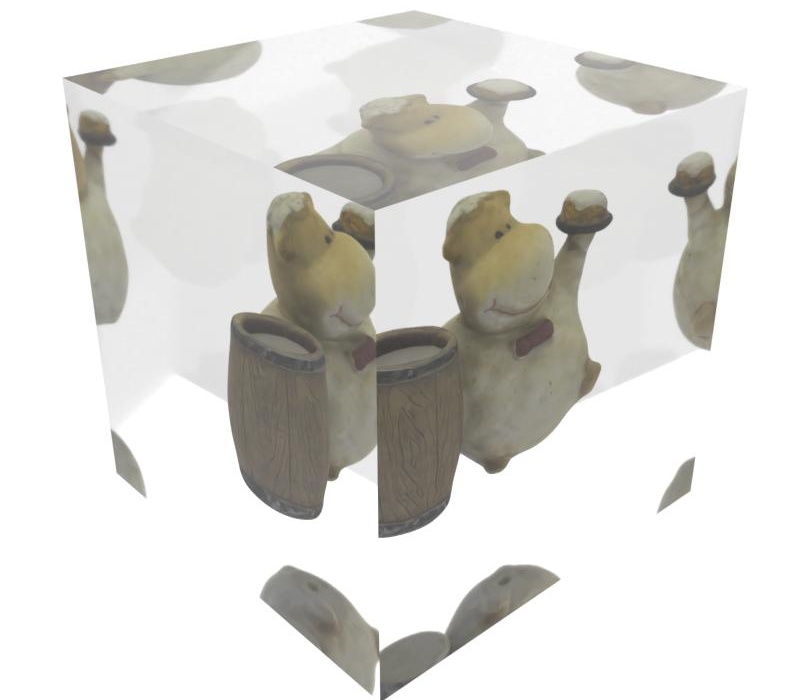}
    \caption*{GT view \\ Chamfer-$L_1\downarrow$}
    \label{fig:ablation_a}
    \end{subfigure}
    \begin{subfigure}[b]{0.23\linewidth}
    \includegraphics[width=1\linewidth]{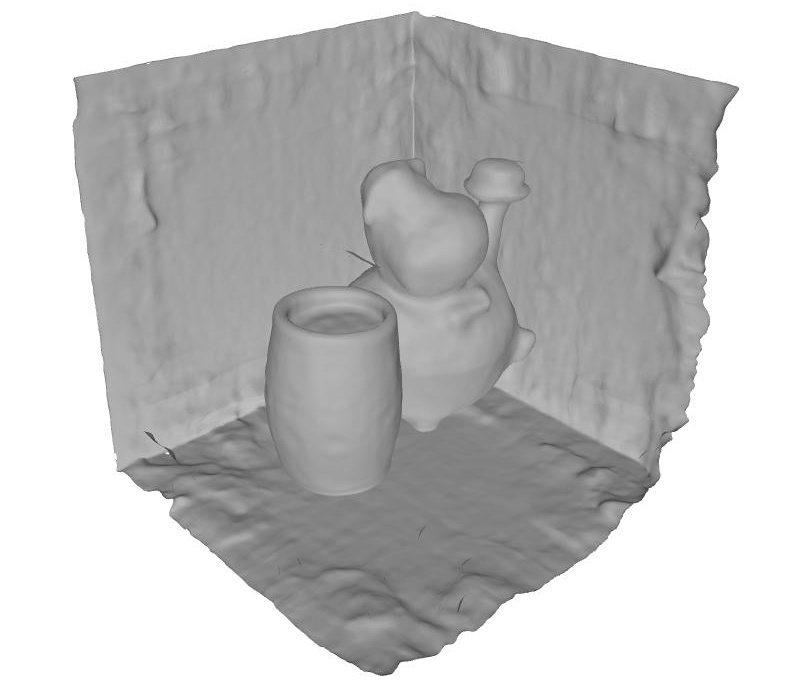}
    \caption*{$\text{NeuS}^+$ \\ $25.21$}
    \label{fig:ablation_b}
    \end{subfigure}
    \begin{subfigure}[b]{0.23\linewidth}
    \includegraphics[width=1\linewidth]{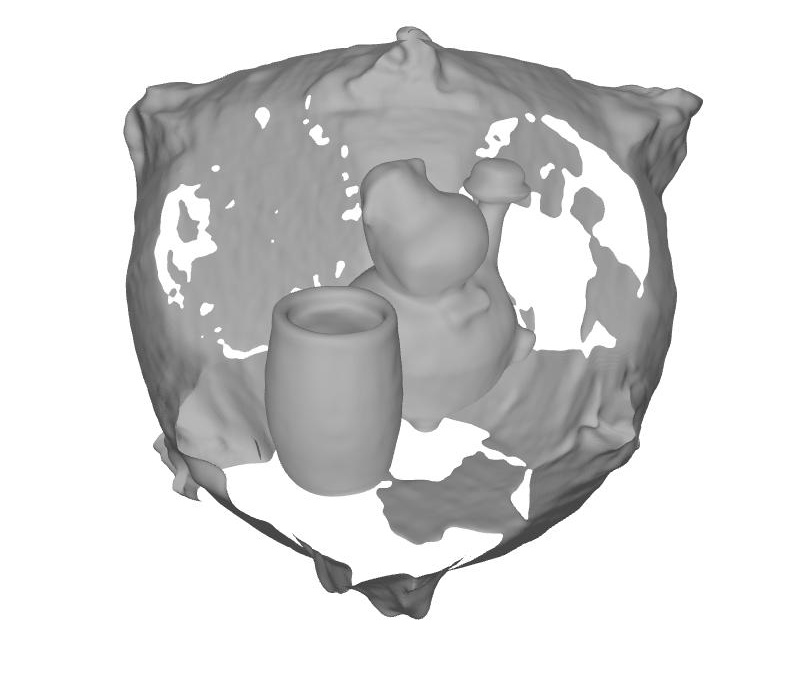}
    \caption*{Ours w/o Trans Loss \\ $21.12$}
    \label{fig:ablation_d}
    \end{subfigure}
    \begin{subfigure}[b]{0.23\linewidth}
    \includegraphics[width=1\linewidth]{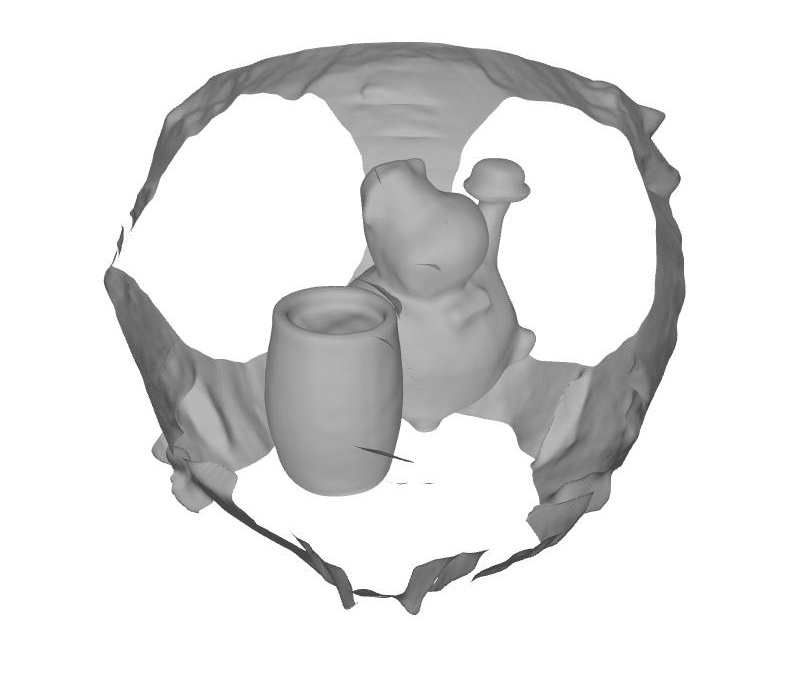}
    \caption*{\textbf{Ours full} \\ $16.80$}
    \label{fig:ablation_e}
    \end{subfigure}
    \caption{Visualization on ablation study.}
    \label{fig:ablation}
    \vspace{-1em}
\end{figure*}

\begin{figure}[ht]
\setlength{\abovecaptionskip}{0pt}
\setlength{\belowcaptionskip}{0pt}
    \centering
    \begin{minipage}{0.85\linewidth}
    \centering
    \begin{subfigure}[b]{0.4\linewidth}
    \centering
    \begin{minipage}{1\linewidth}
        \includegraphics[width=\linewidth, height=\linewidth]{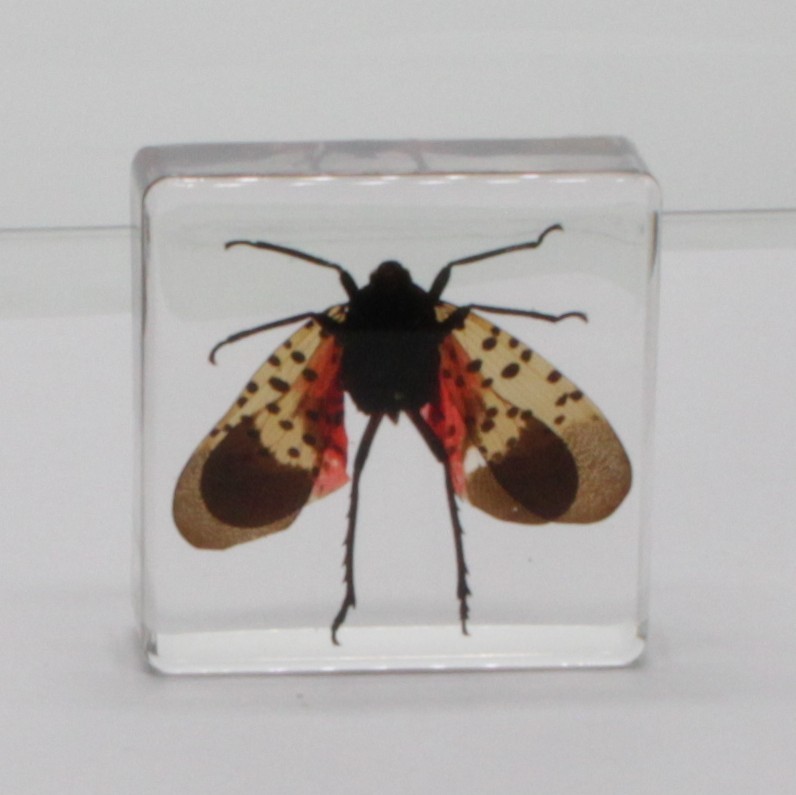}
        \includegraphics[width=\linewidth, height=\linewidth]{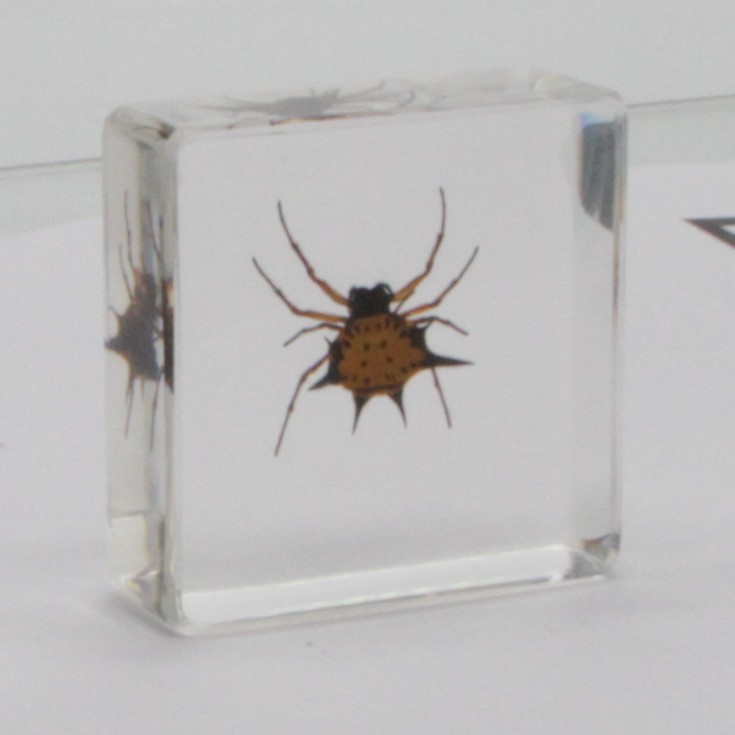}
        \includegraphics[width=\linewidth, height=\linewidth]{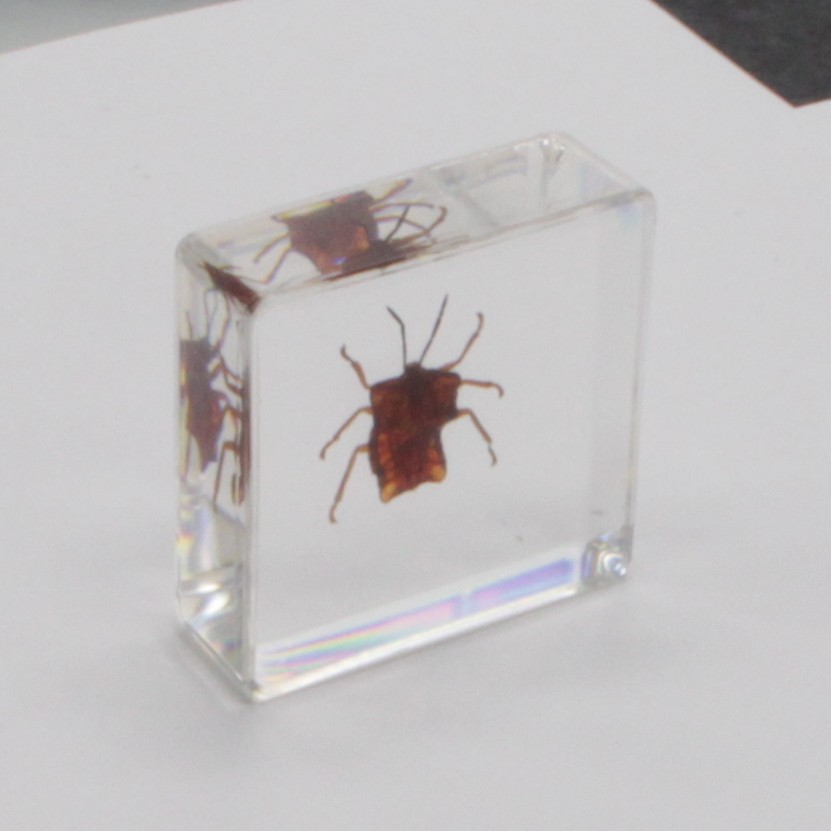}
        \includegraphics[width=\linewidth, height=\linewidth]{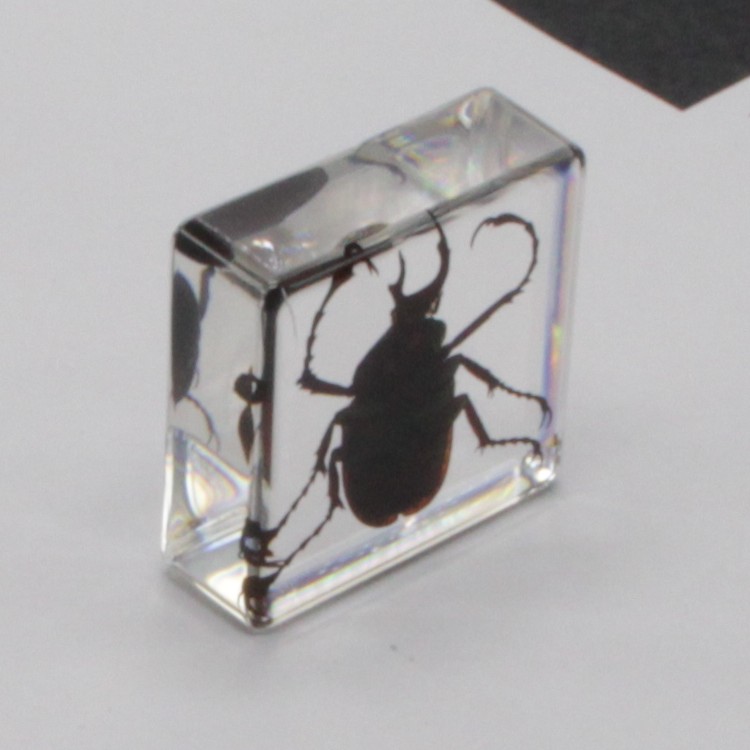}
    \end{minipage}
    \centerline{\small Reference}
    \caption*{}
    \end{subfigure}%
    \hspace{1em}
    \begin{subfigure}[b]{0.4\linewidth}
    \centering
    \begin{minipage}{1\linewidth}
        \includegraphics[width=\linewidth, height=\linewidth]{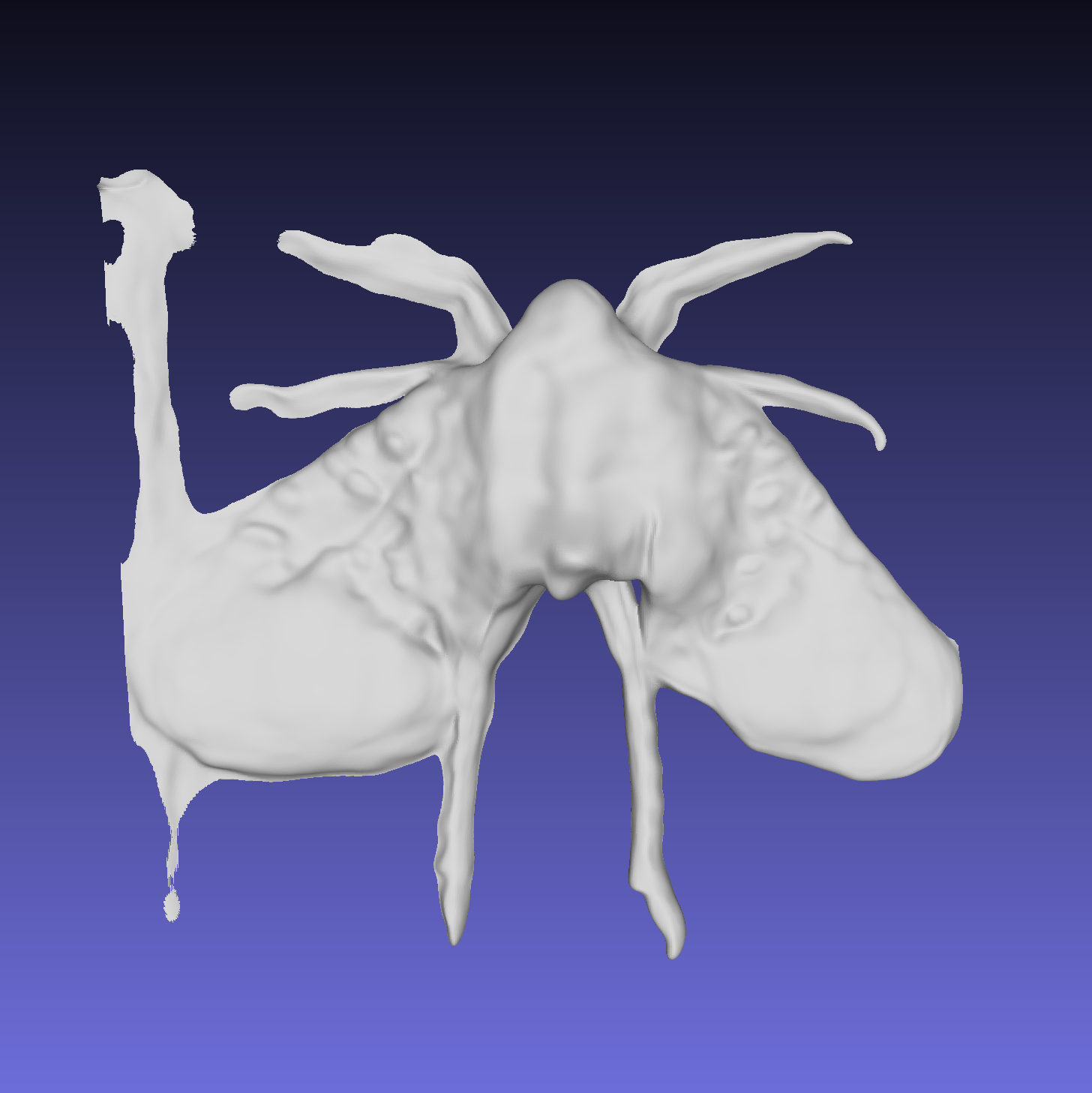}        
        \includegraphics[width=\linewidth, height=\linewidth]{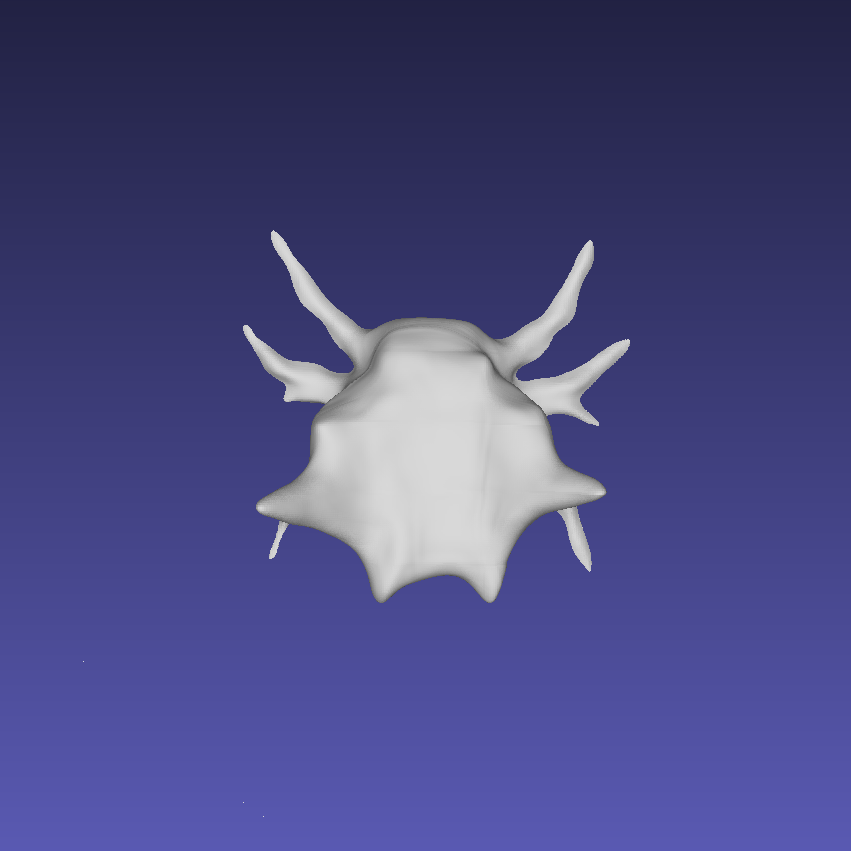}    
        \includegraphics[width=\linewidth, height=\linewidth]{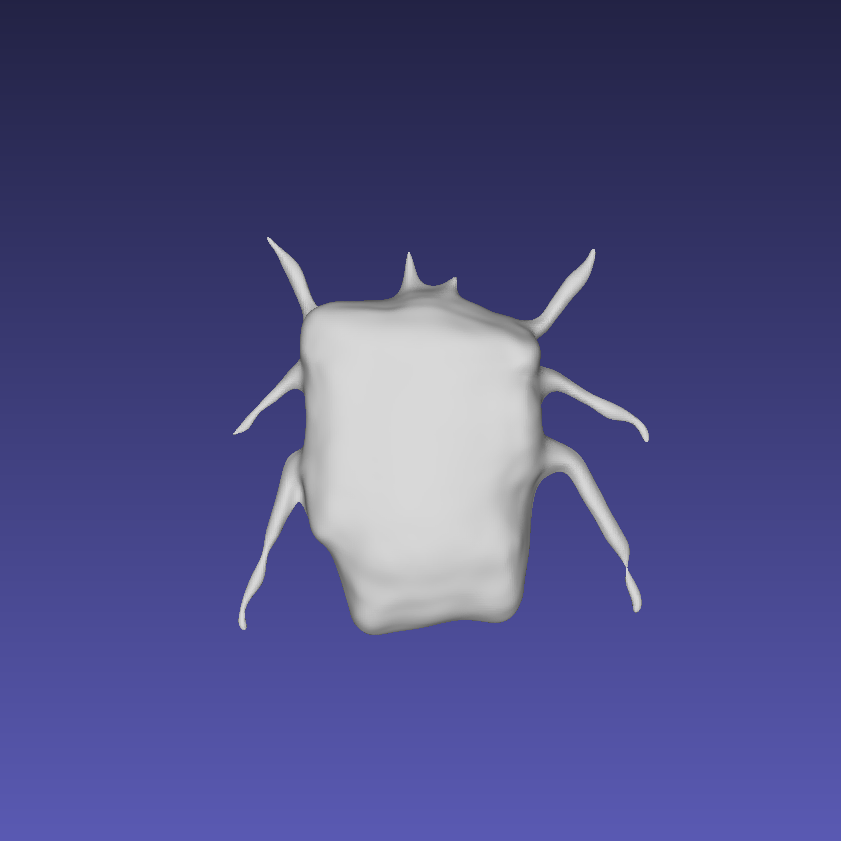}
        \includegraphics[width=\linewidth, height=\linewidth]{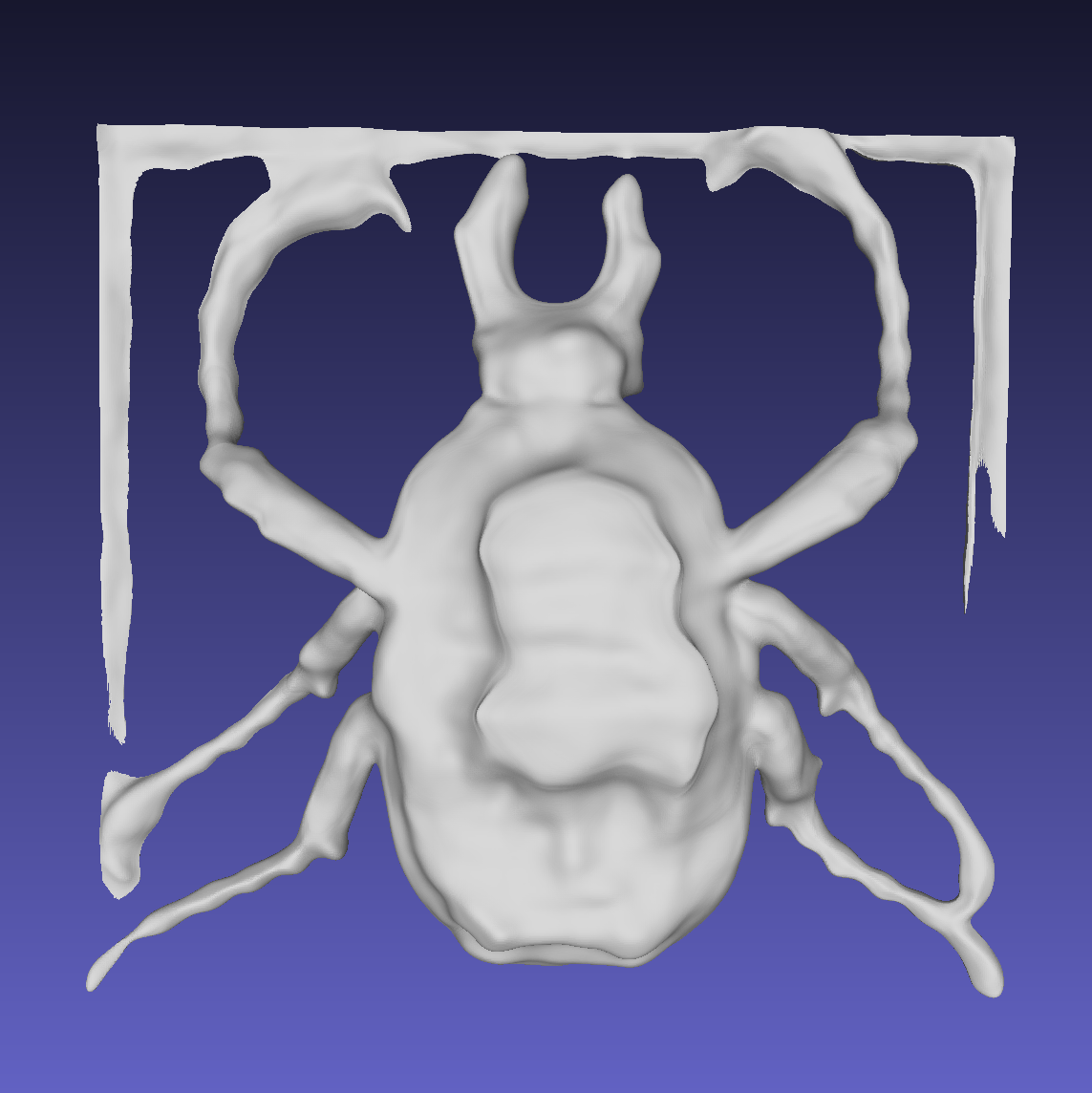}
    \end{minipage}
    \centerline{\small Results}
    \caption*{}
    \end{subfigure}
    \end{minipage}
    \caption{Visualiztion of 3D reconstruction on real dataset.}
    \label{fig:qualitative_real}
    \vspace{-1.5em}
\end{figure}

\subsection{Visualization on real dataset.}
We run ReNeuS and other baseline methods on the real dataset for quantitative comparison.
For IDR, we run it with the mask of the transparent box since we don't have an object mask.
IDR failed to reconstruct anything but a convex hull of the masks. 
Experiment results are shown in \cref{fig:qualitative_real} as well as the supplementary material. 
From the visualization results, we argue that NeReuS performs considerably well on the proposed task. 
Our method can even reconstruct the wings and antennae of small insects which is indeed a difficult problem in normal 3D reconstruction. 

\subsection{Ablation study}
\label{sec:ablation}
\begin{table}[bt]
\resizebox{0.95\linewidth}{!}{
\begin{centering}
\begin{tabular}{lccc}
\toprule   
\vspace{-0.95pt}
\   & $NeuS^+$ & Ours (w/o Trans Loss) & Ours \\ 
\cmidrule{1-4}
\multirow{1}{*}{Chamfer-$L_1\downarrow$} & $22.07$ & $19.08$& $ \mathbf{17.90}$ \\
\multirow{1}{*}{Failure case} & $3$ & $2$ & $\mathbf{0}$ \\ 
\bottomrule 
\end{tabular}
\end{centering}
}
\setlength{\belowcaptionskip}{0cm}
\caption{Results of ablation study.}
\label{tab:ablation_new}
\vspace{-1em}
\end{table}
We conduct ablation studies to verify the effectiveness of our ReNeuS framework.

\textbf{Rendering strategy.} 
We believe that our novel hybrid rendering strategy is the key to solving the problem. 
To verify that, we propose a naive solution based on NeuS \cite{wang2021NeuSLearningNeural} that we only refract the rays once on the space interface.
We refer to it as $\text{NeuS}^+$.
As the result shows in \cref{fig:ablation}, the result of $\text{NeuS}^+$ is actually fully surrounded by another box-shaped mesh.
The reason is that the $\text{NeuS}^+$ method doesn't model the scene with light interaction exactly. 
While training with photometric loss, the network tries to render an image of the statuette on each side, thus producing these fake boundaries. 
As our method takes good care of the light interaction, the fake boundaries are greatly relieved.
Based on this observation, we abandon the post-processing in the ablation study to better evaluate the influence of such errors.
The results are shown in \cref{tab:ablation_new} where \emph{Failure case} means that we fail to retrieve a mesh from the geometric MLP, after trying 3 times.
Apart from accuracy, our method performs consistently in different cases. 
Especially for the \emph{dinosaur} case which is quite challenging due to its thin structure and texture-less appearance, all the two methods failed except the full model.

\textbf{Transmittance loss.}
We also do an ablation study for our sparsity prior item as shown in \cref{fig:ablation}. 
Those dummy facets are dramatically suppressed with the regularization. 

\section{Limitations and Conclusion}
\label{sec:conclusion}
\textbf{Limitation.} The major limitation of our method is the assumption of known geometry and homogeneous background lighting.
The known geometry assumption may be violated due to irregularly shaped transparent containers with rounded edges and corners and the lighting assumption may impede the data collection process.
We notice from the results on real dataset that the violation of the assumptions can lead to a performance drop.
We believe tackling these limitations by jointly reconstructing the object and container as well as modeling the lighting conditions can lead to promising future research directions. 

\textbf{Conclusion.} In this paper, we have defined a new research problem of 3D reconstruction of specimens enclosed in transparent containers raised from the digitization of fragile museum collections.
The image distortion caused by complex multi-path light interactions across interfaces, that separate multiple mediums, poses challenges to recovering accurate 3D geometry of the specimen. 
To tackle the problem, we propose a neural reconstruction method ReNeuS in which we use a novel hybrid rendering strategy that takes into account multiple light reflections and refractions across glass/air interfaces to accurately reconstruct 3D models.
Experiments on both synthetic and our newly captured real dataset demonstrate the efficacy of our method.
It is capable of producing high-quality 3D reconstructions when existing methods fail.

\textbf{Acknowledgements.} The project is supported by CSIRO Julius Career Award, CSIRO Early Research Career (CERC) Fellowship and CSIRO Future Digital Manufacturing Fund (FDMF) Postdoc Fellowship. Jinguang is supported by CSIRO Data61 PhD Scholarship.


{\small
\bibliographystyle{ieee_fullname}
\bibliography{egbib}
}

\end{document}